\def\year{2019}\relax
\documentclass[letterpaper]{article} 
\usepackage{aaai19}  
\usepackage{times}  
\usepackage{helvet}  
\usepackage{courier}  
\usepackage{url}  
\usepackage{graphicx}  
\usepackage{amsmath}
\usepackage{amsfonts}
\usepackage{amsthm}
\usepackage{xcolor}

\usepackage{algorithm}
\usepackage{algorithmic}

\DeclareMathOperator*{\argmax}{arg\,max}

\newtheorem{thm}{Theorem}

\allowdisplaybreaks

\newcommand{\norm}[1]{\left\lVert#1\right\rVert}

\title{Diverse Exploration via Conjugate Policies for Policy Gradient Methods}

\author{Andrew Cohen \\ Binghamton University  \\acohen13@binghamton.edu \And 
    Xingye Qiao \\ Binghamton University \\ qiao@math.binghamton.edu \And 
    Lei Yu \\ Binghamton University \\ Yantai University \\ lyu@cs.binghamton.edu \AND
    Elliot Way \\ Binghamton University \\ eway1@binghamton.edu \And
    Xiangrong Tong \\ Yantai University \\ txr@ytu.edu.cn}
\begin{document}

\maketitle

\begin{abstract}
We address the challenge of effective exploration while maintaining good performance in policy gradient methods. As a solution, we propose diverse exploration (DE) via conjugate policies. DE learns and deploys a set of conjugate policies which can be conveniently generated as a byproduct of conjugate gradient descent. We provide both theoretical and empirical results showing the effectiveness of DE at achieving exploration, improving policy performance, and the advantage of DE over exploration by random policy perturbations. 
\end{abstract}

\section{Introduction}
Policy gradient (PG) \cite{NAC,TRPO,PG,acktr} methods in reinforcement learning (RL)~\cite{sutton} have shown the ability to train large function approximators with many parameters but suffer from slow convergence and data inefficiency due to a lack of exploration. Achieving exploration while maintaining effective operations is a challenging problem as exploratory decisions may degrade performance. Conventional exploration strategies which achieve exploration via noisy action selection (like $\epsilon$-greedy~\cite{sutton}) or actively reducing uncertainty (like R-MAX~\cite{Braf-Tenn03}) do not guarantee performance of the behavior policy.

This work follows an alternative line that performs exploration in {\it policy space}, and is inspired by two recent advances: Diverse Exploration (DE)~\cite{DEFSI} and parameter space noise for exploration~\cite{PSNoise}. The key insight of DE is that, in many domains, there exist multiple different policies at various levels of policy quality. Effective exploration can be achieved without sacrificing exploitation if an agent learns and deploys a set of diverse behavior policies within some policy performance constraint. This work shares a similar intuitive motivation in the context of PG methods: multiple parameterizations of ``good'' but, importantly, {\it different} policies exist in the local region of a main policy. Deploying a set of these policies increases the knowledge of the local region and can improve the gradient estimate in policy updates. Though similarly motivated, this work provides distinct theoretical results and an algorithmic solution to a unique challenge in PG methods: {\it to maximally explore local policy space in order to improve the gradient estimate while ensuring performance}.

Parameter space noise for exploration~\cite{PSNoise} can be thought of as a DE approach specific to the PG context. To achieve exploration, different behavior policies are generated by randomly perturbing policy parameters. To maintain the guarantees of the policy improvement step from the previous iteration, the magnitude of these perturbations has to be limited which inherently limits exploration.
Thus, for effective exploration in PG methods, we need an optimal diversity objective and a principled approach of maximizing diversity. In light of this, we propose DE by  {\it conjugate policies} that maximize a theoretically justified Kullback\textendash Leibler (KL) divergence objective for exploration in PG methods.

The novel contributions of this paper are three--fold. {\it First}, it proposes a DE solution via conjugate policies for natural policy gradient (NPG) methods. DE learns and deploys a set of conjugate policies in the local region of policy space and follows the natural gradient descent direction during each policy improvement iteration. 

{\it Second}, it provides formal explanation for why DE via conjugate policies is effective in NPG methods. Our theoretical results show that: (1) maximizing the diversity (in terms of KL divergence) among perturbed policies is inversely related to the variance of the perturbed gradient estimate, contributing to more accurate policy updates; and (2) conjugate policies generated by conjugate vectors maximize pairwise KL divergence among a constrained number of perturbations. In addition to justifying DE via conjugate policies, these theoretical results explain why parameter space noise~\cite{PSNoise} improves upon NPG methods but is not optimal in terms of the maximum diversity objective proposed in this work. 

{\it Finally}, it develops a general algorithmic framework of DE via conjugate policies for NPG methods. The algorithm efficiently generates conjugate policies by taking advantage of conjugate vectors produced in each policy improvement iteration when computing the natural gradient descent direction. Experimental results based on Trust Region Policy Optimization (TRPO) \cite{TRPO} on three continuous control domains show that TRPO with DE significantly outperforms the baseline TRPO as well as TRPO with random perturbations. 

\section{Preliminaries}\label{prelim}
RL problems are described by Markov Decision Processes (MDP)~\cite{puterman1994markov}. An MDP, $M$, is defined as a 5-tuple, $M =(S,A,P,\mathcal{R},\gamma)$, where $S$ is a fully observable set of states, $A$ is a set of possible actions, $P$ is the state transition model such that $P(s'|s,a) \in [0,1]$ describes the probability of transitioning to state $s'$ after taking action $a$ in state $s$, $\mathcal{R}_{s,s'}^a$ is the expected value of the immediate reward $r$ after taking $a$ in $s$, resulting in $s'$, and $\gamma\in (0,1)$ is the discount factor on future rewards. A {\it trajectory} of length $T$ is an ordered set of transitions: $\tau= \{s_0,a_0,r_1,s_1,a_1,r_2,...,s_{T-1},a_{T-1},r_{T}\}$.

A solution to an MDP is a policy, $\pi(a|s)$ which provides the probability of taking action $a$ in state $s$ when following policy $\pi$.
The performance of policy $\pi$ is the expected discounted return
\begin{align*}
J(\pi) = \mathbb{E}_{\tau}[R(\tau)] = \mathbb{E}_{s_0,a_0..}[\sum_{t=0}^\infty \gamma^tr(a_t,s_t)]\\
where\ s_0 \sim \rho(s_0),\ a_t \sim \pi( \cdot | s_t),\ s_{t+1} \sim P(\cdot | s_t,a_t)
\end{align*}
and $\rho(s_0)$ is the distribution over start states.

The state-action value function, value function and advantage function are defined as:
\begin{align*}
Q_{\pi}(s_t,a_t) = \mathbb{E}_{s_{t+1},a_{t+1}..}[\sum_{l=0}^\infty \gamma^lr(a_{t+l},s_{t+l})]\\
V_{\pi}(s_t) = \mathbb{E}_{a_t,s_{t+1}..}[\sum_{l=0}^\infty \gamma^lr(a_{t+l},s_{t+l})]\\
A_{\pi}(s_t,a_t) = Q_{\pi}(s_t,a_t)- V_{\pi}(s_t)\\ 
where\ a_t \sim \pi( \cdot | s_t),\ s_{t+1} \sim P(\cdot | s_t,a_t).
\end{align*}

In policy gradient methods, $\pi_{\theta}$ is represented by a function approximator such as a neural network parameterized by vector $\theta$. These methods maximize via gradient descent on $\theta$ the expected return of $\pi_{\theta}$ captured by the objective function:
\begin{equation*}
\max_{\theta}J(\theta)= \mathbb{E}_{\tau}[R(\tau)].
\end{equation*}
The gradient of the objective $J(\theta)$ is:
\begin{equation*}
\nabla_{\theta} J(\theta)=\mathbb{E}_{\tau}[\sum_{t=0}^T \nabla_{\theta}\log(\pi(a_t | s_t;\theta))R_t(\tau)],
\end{equation*}
which is derived using the likelihood ratio. 
This is estimated empirically via
\begin{equation*}
\nabla_{\theta} \hat{J}(\theta)=\frac{1}{N}\sum_{i=0}^N[\sum_{t=0}^T \nabla_{\theta}\log(\pi(a_t | s_t;\theta))\hat{A}_{\pi}(s_t, a_t)],
\end{equation*}
where an empirical estimate of the advantage function is used instead of $R_t(\tau)$ to reduce variance and $N$ is the number of trajectories.
The policy update is then $\theta_{i+1} = \theta_i + \alpha \nabla_{\theta} \hat{J}(\theta)$ where $\alpha$ is the stepsize. This is known as the `vanilla' policy gradient.

\subsection{Natural Gradient Descent and TRPO}
A shortcoming of vanilla PG methods is that they are not invariant to the scale of parameterization nor do they consider the more complex manifold structure of parameter space.  Natural Gradient Descent methods \cite{NGD,InfoGeom} address this by correcting for the curvature of the parameter space manifold by scaling the gradient with the inverse Fisher Information Matrix (FIM) $F_{\theta}$ where

\begin{equation*}
F_{ij,\theta} = -\mathbb{E}_{s \sim \rho}[\frac{\partial}{\partial \theta_i}\frac{\partial}{\partial \theta_j}\log(\pi_{\theta}(\cdot |s))]
\end{equation*}
is the $i,j$th entry in the FIM and $\rho$ is the state distribution induced by policy $\pi_\theta$.
The natural policy gradient descent direction and policy update is then
\begin{align*}
\nabla_{\theta} \tilde{J}(\theta)= F_{\theta}^{-1}\nabla_{\theta} J(\theta),\ 
\theta_{i+1} = \theta_i + \alpha \nabla_{\theta} \tilde{J}(\theta).
\end{align*}

Selecting the stepsize $\alpha$ is not trivial. TRPO \cite{TRPO}, a robust and state of the art approach, follows the natural gradient descent direction from the current policy $\pi_{\theta'}$ but enforces a strict KL divergence constraint by optimizing
\begin{align*}
\max_{\theta} \mathbb{E}_{s \sim \rho_{\theta'},a \sim \pi_{\theta'}}[\frac{\pi(a|s;\theta)}{\pi(a|s;\theta')}R_t(\tau)]\\
subject\ to\ D_{KL}(\pi_{\theta}||\pi_{\theta'}) \le \delta
\end{align*}
which is equivalent to the standard objective.
The KL divergence between two policies is 
\begin{equation*}
D_{KL}(\pi_{\theta}||\pi_{\theta'}) := \mathbb{E}_{s\sim \rho}[D_{KL}(\pi_{\theta}(\cdot | s)||\pi_{\theta'}(\cdot | s))].
\end{equation*}
Via a Taylor expansion of $D_{KL}$, one can obtain the following local approximation
\begin{align*}
\tilde{D}_{KL}(\theta||\theta+d\delta) = \frac{1}{2}(\theta + d\delta - \theta)^T F_{\theta} (\theta + d\delta - \theta)\\
=\frac{1}{2}{d\delta}^T F_{\theta} d\delta,
\end{align*}
where $\theta + d\delta$ and $\theta$ parameterize two policies.
This approximation is used throughout this work.

\subsection{Parameter Space Noise for Exploration}

The reinforcement learning gradient estimation can be generalized with inspiration from evolutionary strategies \cite{NES} by sampling parameters $\theta$ from a search distribution  $\mathcal{N}(\phi,\Sigma)$ \cite{PSNoise}.
\begin{multline}\label{congrad}
\nabla_{\phi,\Sigma} \mathbb{E}_{\substack{\theta \sim \mathcal{N}(\phi,\Sigma)\\ \tau \sim \pi}}[R(\tau)]=\\
\mathbb{E}_{\substack{\epsilon \sim \mathcal{N}(0,I)\\ \tau \sim \pi}}[\sum_{t=0}^T \nabla_{\phi,\Sigma}\log(\pi(a_t | s_t;\phi + \epsilon \Sigma^\frac{1}{2}))R_t(\tau)]
\end{multline}
which is derived using likelihood ratios and reparameterization \cite{RepTrick}. 
The corresponding empirical estimate is
\begin{equation*}
\frac{1}{N}\sum_{i=0}^N[\sum_{t=0}^T \nabla_{\phi,\Sigma}\log(\pi(a_t | s_t;\phi + \epsilon_i \Sigma^\frac{1}{2}))\hat{A}_{\pi}(s_t, a_t)].
\end{equation*}
This gradient enables exploration because it aggregates samples from multiple policies (each $\epsilon_i$ defines a different, perturbed policy). It may seem that using trajectories collected from perturbed policies introduces off-policy bias (and it would for the standard on-policy gradient estimate). However, the generalized objective in Equation~(\ref{congrad}) does not have this issue since the gradient is computed over a perturbation distribution. 

Perturbation approaches suffer from an exploration-exploitation dilemma. Large perturbations increase exploration but potentially degrade performance since the perturbed policy becomes significantly different from the main, unperturbed policy. A small perturbation provides limited exploration but will benefit from online performance similar to that of the main policy.  Our approach, DE, is designed to maximize diversity in a limited number of perturbations within a bounded local region of parameter space to address this tradeoff which random perturbations do not.  From here on, we refer to Equation~(\ref{congrad}) as the ``perturbed gradient estimate'' and refer to the approach that samples random perturbations as RP.
\section{Theoretical Analysis}\label{theory}
In this section, we first prove that increasing the KL divergence between perturbed policies reduces the variance of the  perturbed gradient estimate. Further, we prove that conjugate vectors maximize pairwise KL divergence among a constrained number of perturbations. 

\subsection{Variance Reduction of Parameter Perturbation Gradient Estimator}
We consider the general case where $\epsilon \sim \mathbb{P}$ where $\mathbb{P}$ is the perturbation distribution.  When $\mathbb{P} = \mathcal{N}(0,\Sigma)$, we recover the gradient in Equation~(\ref{congrad}) in the Preliminaries. To simplify notations in the variance analysis of the perturbed gradient estimate, we write $\epsilon$ as shorthand for $\phi + \epsilon$ and let $\pi_{\epsilon}$ be the policy with parameters $\phi$ perturbed by $\epsilon$. Moreover,
\begin{align*}
G_{\epsilon} := \mathbb{E}_{\tau \sim \pi_{\epsilon}}[\sum_{t=0}^T \nabla_{\phi}\log(\pi_{\epsilon}(a_t | s_t))R_t(\tau)]
\end{align*}
is the gradient with respect to $\phi$ with perturbation $\epsilon$. The final estimate to the true gradient in Equation 
~(\ref{congrad}) is the Monte Carlo estimate of $G_{\epsilon_i}$ ($1\le i\le k$) over $k$ perturbations. For any $\epsilon_i$, $G_{\epsilon_i}$ is an unbiased estimate of the gradient so the averaged estimator is too. Therefore, by reducing the variance, we reduce the estimate's mean squared error. 
The variance of the estimate over $k$ perturbations $\epsilon_i$ is
\begin{multline}\label{variance}
\mathbb{V}(\frac{1}{k}\sum_{i=1}^kG_{\epsilon_i}) =\\
\frac{1}{k^2}\sum_{i=1}^k \mathbb{V}_{\epsilon_i}(G_{\epsilon_i})+\frac{2}{k^2}\sum_{i=1}^{k-1}\sum_{j=i+1}^kCov_{\epsilon_i,\epsilon_j}(G_{\epsilon_i},G_{\epsilon_j})
\end{multline}
where $\mathbb{V}_{\epsilon_i}(G_{\epsilon_i})$ is the variance of the gradient estimate $G_{\epsilon_i}$and $Cov_{\epsilon_i,\epsilon_j}(G_{\epsilon_i},G_{\epsilon_j})$ is the covariance between the gradients $G_{\epsilon_i}$ and $G_{\epsilon_j}$.

$\mathbb{V}_{\epsilon_i}(G_{\epsilon_i})$ is equal to a constant for all $i$ because $G_{\epsilon_i}$ are identically distributed. So, the first term in Equation~(\ref{variance}) approaches zero as $k$ increases and does not contribute to the asymptotic variance. The covariance term determines whether the overall variance can be reduced. 
To see this, consider the extreme case when $G_{\epsilon_i}=G_{\epsilon_j}$ for $i\neq j$. Equation~(\ref{variance}) becomes $\mathbb{V}(\frac{1}{k}\sum_{i=1}^kG_{\epsilon_i})=\mathbb{V}_{\epsilon_1}(G_{\epsilon_1})$ because all $Cov_{\epsilon_i,\epsilon_j}(G_{\epsilon_i},G_{\epsilon_j}) = \mathbb{V}_{\epsilon_1}(G_{\epsilon_1})$. The standard PG estimation (i.e. TRPO) falls into this extreme as a special case of the perturbed gradient estimate where all perturbations are the zero vector.

Next consider the special case where $Cov_{\epsilon_i,\epsilon_j}(G_{\epsilon_i},G_{\epsilon_j})=0$ for $i \neq j$. Then, the second term vanishes and $\mathbb{V}(\frac{1}{k}\sum_{i=1}^kG_{\epsilon_i})=O(k^{-1})$. The RP approach strives for this case by  i.i.d. sampling of perturbations $\epsilon$. This explains why RP was shown to outperform TRPO in some experiments \cite{PSNoise}.  However, it is important to note that i.i.d. $\epsilon$ do not necessarily produce uncorrelated gradients $G_{\epsilon}$ as this depends on the local curvature of the objective function. For example, perturbations in a flat portion of parameter space will produce equal gradient estimates that are perfectly positively correlated. Thus, $G_{\epsilon_i}$ are identically distributed but not necessarily independent. This suggests that using a perturbation distribution such as $\mathcal{N}(0,\Sigma)$ may suffer from potentially high variance if further care is not taken. This work develops a principled way to select perturbations in order to reduce the covariance.

There are two major sources of variance in the covariance terms; the correlations among $\nabla_{\phi}\log(\pi_{\epsilon_i})$ and $\nabla_{\phi}\log(\pi_{\epsilon_j})$ and correlations related to $R_t(\tau)$. The difference in performance of two policies (as measured by $R_t(\tau)$) can be bounded by a function of the average KL divergence between them \cite{TRPO}. So, the contribution to the covariance from $R_t(\tau)$ will be relatively fixed since all perturbations have a bounded KL divergence to the main policy. In view of this, we focus on controlling the correlation between $\nabla_{\phi}\log(\pi_{\epsilon_i})$ and $\nabla_{\phi}\log(\pi_{\epsilon_j})$.

This brings us to Theorem~\ref{min_variance} (with proof in the supplementary file)
in which we show that maximizing the diversity in terms of KL divergence between two policies $\pi_{\epsilon_i}$ and $\pi_{\epsilon_j}$ minimizes the trace of the covariance between $\nabla_{\phi}\log(\pi_{\epsilon_i})$ and $\nabla_{\phi}\log(\pi_{\epsilon_j})$.

\begin{thm}\label{min_variance}
Let $\epsilon_i$ and $\epsilon_j$ be two perturbations such that $\norm{\epsilon_i}_2$ = $\norm{\epsilon_j}_2=\delta_\epsilon$. Then, (1) the trace of $\textrm{Cov}(\nabla_{\phi}\log(\pi_{\epsilon_j}),\nabla_{\phi}\log(\pi_{\epsilon_i}))$ is minimized and (2) $\frac{1}{2}(\epsilon_j-\epsilon_i)^T\hat F(\epsilon_i)(\epsilon_j-\epsilon_i)$ the estimated KL divergence $D_{KL}(\pi_{\epsilon_i}||\pi_{\epsilon_j})$ is maximized, when $\epsilon_i=-\epsilon_j$ and they are along the direction of the eigenvector of $F(\epsilon_i)$ with the largest eigenvalue.
\end{thm}
This theorem shows that, when two perturbations $\epsilon_i$ and $\epsilon_j$ have a fixed $L2$ norm $\delta_\epsilon$, the perturbations that maximize the KL divergence $D_{KL}(\pi_{\epsilon_i}||\pi_{\epsilon_j})$ and also minimize the trace of the covariance $\textrm{Cov}(\nabla_{\phi}\log(\pi_{\epsilon_j}),\nabla_{\phi}\log(\pi_{\epsilon_i}))$ are uniquely defined by the positive and negative directions along the eigenvector with the largest eigenvalue. This provides a principled way to select two perturbations to minimize the covariance. 

\subsection{Conjugate Vectors Maximize KL Divergence}

In domains with high sample cost, there is likely a limit on the number of samples an agent can collect and so too on the number of policies which can be deployed per iteration.  Therefore, it is important to generate a small number of perturbations which yield maximum variance reduction. Theorem 1 shows that the reduction of the covariance can be done by maximizing the KL divergence. We show in the theorem that eigenvectors can achieve this. Eigenvectors are a special case of what are known as conjugate vectors. Later in this section, Theorem 2 shows that when there is a fixed set of $k$ perturbations, conjugate vectors maximize the sum of the pairwise KL divergences. We first establish notation. 

Since the FIM $F_{\phi}$ is symmetric positive definite, there exist $n$ conjugate vectors $\mathcal{U} = \{\mu_1,\mu_2,\ .\ .\ ,\mu_n\}$ with respect to $F_{\phi}$ where $n$ is the length of the parameter vector $\phi$. Formally, $\mu_i$ and $\mu_j,\ i \ne j$ are conjugate if $\mu_i^T F_{\phi} \mu_j=0$.  We define
$\pi_i$ and $\pi_j$ as {\it conjugate policies} if their parameterizations can be written as $\phi + \mu_i$ and $\phi + \mu_j$ for two conjugate vectors $\mu_i$ and $\mu_j$.
$\mathcal{U}$ forms a basis for $\mathbb{R}^n$ so any local perturbation $\epsilon$ to $\phi$, after scaling, can be written as a linear combination of $\mathcal{U}$,
\begin{align}
\epsilon = \eta_1\mu_1 + \eta_2\mu_2 +\ .\ .\ +\eta_n\mu_n\ 
where\ \norm{\eta}\le 1,
\end{align}
For convenience, we assume that $\eta_i \ge 0$. Since the negative of a conjugate vector is also conjugate, if there is a negative $\eta_i$, we may flip the sign of the corresponding $\mu_i$ to make it positive.

Recall the approximation of KL divergence from the Preliminaries,
$$
\tilde{D}_{KL}(\phi||\phi+\epsilon) =
\frac{1}{2}{\epsilon}^T F_{\phi} \epsilon
$$
The measure of KL divergence that concerns us is the total divergence between all {\it pairs} of perturbed policies:
\begin{multline}\label{pairwise}
\sum_{i=1}^{k-1} \sum_{j=i+1}^k \tilde{D}_{KL}(\phi+\epsilon_j||\phi+\epsilon_i)=\\
\sum_{i=1}^{k-1} \sum_{j=i+1}^k \frac{1}{2}(\epsilon_i-\epsilon_j)^TF_{\phi}(\epsilon_i-\epsilon_j)
\end{multline}
where $k$ is the number of perturbations. Note that we use $\phi$ and not $\phi + \epsilon$ in the subscript of the FIM which would be more precise with respect to the local approximation. The use of the former is a practical choice which allows us to estimate a single FIM and avoid estimating the FIM of each perturbation.  Estimating the FIM is already a computational burden and, since perturbations are small and bounded, using $F_\phi$ instead of $F_{\phi+\epsilon}$ has little effect and performs well in practice as demonstrated in experiments.   For the remainder of this section, we omit $\phi$ in the subscript of $F$ for convenience. The constraint on the number of perturbations brings us to the following optimization problem that optimizes a set of perturbations $\mathcal{P}$ to maximize~(\ref{pairwise}) while constraining $|\mathcal{P}|$.
\begin{multline}\label{objf}
\begin{aligned}
\mathcal{P}^* =\argmax_{\mathcal{P}}\sum_{i=1}^{k-1}\sum_{j=i+1}^k\tilde{D}_{KL}(\phi + \epsilon_j||\phi + \epsilon_i)\\
subject\ to\ |\mathcal{P}| = k \le n
\end{aligned}
\end{multline}

We define $\|\cdot\|_F$ as the norm induced by $F$, that is, $$\|x\|_F=x^TFx.$$

Without the loss of generality, assume the conjugate vectors are ordered with respect to the $F$-norm,
$$
\norm{\mu_1}_F \ge \norm{\mu_2}_F \ge\ .\ .\ \ge \norm{\mu_n}_F.
$$

The following theorem gives an optimal solution to the objective~(\ref{objf}). The proof comes easily by induction on $k$ (full details in the supplementary file).
\begin{thm}\label{opt}
The set of conjugate vectors $\{\mu_1, \mu_2,\ .\ .\ ,\mu_k\}$ maximize the objective~(\ref{objf}) among any $k$ perturbations.
\end{thm}

If we relax the assumption that $\eta_i \ge 0$, then the set of vectors that maximize the objective~(\ref{objf}) simply includes the negative of each conjugate vector as well, i.e., $\mathcal{P} = \{\mu_1, -\mu_1, \mu_2, -\mu_2\ .\ .\ ,\mu_{\frac{k}{2}}, -\mu_{\frac{k}{2}}\}$.  Including the negatives of perturbations is known as  {\it symmetric sampling} \cite{PGPE} which is discussed further in the next section.

Theorem~\ref{opt} makes clear that randomly generated perturbations will be sub-optimal with high probability with respect to the  objective~(\ref{objf}) because the optimal solution is uniquely the top $k$ conjugate vectors. Identifying the top $k$ conjugate vectors in each iteration of policy improvement will require significant computation when the FIM is large. Fortunately, there exist computationally efficient methods of generating sequences of conjugate vectors such as conjugate gradient descent \cite{Nocedal} (to be discussed), although they may not provide the top $k$. From Theorem~\ref{opt}, we also observe that when all conjugate vectors have the same $F$-norm, then any set of $k$ conjugate vectors maximize the objective~(\ref{objf}). If we bound the perturbation radius (the maximum KL divergence a perturbation may have from the main policy) as in \cite{PSNoise}, DE achieves a computationally efficient, optimal solution to the objective~(\ref{objf}).

\section{Method}
In this section, we first discuss an efficient method to generate conjugate policies and then provide a general algorithmic framework of DE via conjugate policies.
\subsection{Generating Conjugate Policies}
Generating conjugate policies by finding the top $k$ conjugate vectors is feasible but computationally expensive. It would require estimating the full empirical FIM of a large neural network (for which efficient approximate methods exist \cite{kfac}) and a decomposition into conjugate vectors. We avoid this additional computational burden altogether and generate conjugate policies by taking advantage of runoff from the conjugate gradient descent (CGD) algorithm \cite{Nocedal}. CGD is often used to efficiently approximate the natural gradient descent direction as in \cite{TRPO}. 

CGD iteratively minimizes the error in the estimate of the natural gradient descent direction along a vector conjugate to all minimized directions in previous iterations.  We utilize these conjugate vectors in DE to be used as perturbations. Although these are not necessarily the top $k$ conjugate vectors, they are computed essentially for free because they are generated from one application of CGD when estimating the natural gradient descent direction. To account for the suboptimality, we introduce a perturbation radius $\delta_p$ such that for any perturbation $\epsilon$ 
\begin{align}
\tilde{D}_{KL}(\phi||\phi+\epsilon) \le \delta_p.
\end{align}
We can perform a line search along each perturbation direction such that $\tilde{D}_{KL}(\phi||\phi+\epsilon) = \delta_p$. With this constraint, the use of any $k$ vectors are optimal as long as they are conjugate and the benefit comes from achieving the optimal {\it pairwise} divergence.

For each conjugate vector, we also include its negative (i.e., symmetric sampling) as motivated by the more general form of Theorem~\ref{opt} with relaxed assumptions (without $\eta_i >0$). In methods following different gradient frameworks, symmetric sampling was used to improve gradient estimations by alleviating a possible bias due to a skewed reward distribution \cite{PGPE}.
Finally, we linearly reduce $\delta_p$ motivated by the observation in \cite{DEFSI} that as a policy approaches optimal there exist fewer policies with similar performance.

\subsection{Algorithm Framework}
\begin{algorithm}
\caption{\textsc{Diverse\_Exploration($\pi_{1}$, $k$, $\beta$, $\beta_k$, $\delta_p$)}}
\label{DPG}
	{\bf Input:} $\pi_{1}$: starting policy, $k$: number of conjugate policies to generate, $\beta$: number of steps to sample from main policy, $\beta_k$: number of steps to sample per conjugate policy, $\delta_p$: perturbation radius
	\begin{algorithmic}[1]
		\STATE Initialize conjugate policies $\mathcal{P}_1$ as $k$ copies of $\pi_{1}$
		\FOR {$i=1,2\ .\ .$}
			\STATE $\mathcal{S}_i \gets$ sample $\beta$ steps from $\pi_i$ and $\beta_k$ steps from each conjugate policy $\pi \in \mathcal{P}_i$ //sample main and diverse policies 
			\STATE $\pi_{i+1}, \mathcal{P}_{i+1} \gets$ $policy\_improvement(\mathcal{S}_i,\ \pi_i,\ k,\ \delta_p)$
		\ENDFOR
	\end{algorithmic}
\end{algorithm}

A general framework for DE is sketched in Algorithm~\ref{DPG}.  In line 1,
DE assumes a starting policy $\pi_{1}$ (e.g., one generated randomly) which is used to initialize conjugate policies as exact copies. The initial parameterization of $\pi_1$ is the mean vector $\phi_1$. The number of conjugate policies to be generated is user defined by an argument $k$. The number of samples to collect from the main and conjugate policies are specified by $\beta$ and $\beta_k$, respectively.  The relative values of $k$, $\beta$ and $\beta_k$ control how much exploration will be performed by conjugate policies. It's worth noting that DE reduces to the standard PG algorithm when $k=0$ or $\beta_k=0$.

In the $i$th iteration, after sampling the main and conjugate policies in line 3, line 4 updates $\phi_i$ via natural gradient descent using the perturbed gradient estimate and returns  the updated policy $\pi_{i+1}$ parameterized by $\phi_{i+1}$ and the set of conjugate policies $\mathcal{P}_{i+1}$ parameterized by $\phi_{i+1}$ perturbed by conjugate vectors; $policy\_improvement$ is a placeholder for any RL algorithm that accomplishes this. Computing perturbations could be done in a separate subroutine (i.e. estimating the FIM and taking an eigendecomposition). When computing the natural gradient by CGD as discussed in the previous section, the intermediate conjugate vectors are saved to be used as perturbations. 

\section{Empirical Study}
We evaluate  the impact of DE via conjugate policies on TRPO \cite{TRPO}. TRPO is state-of-the-art in its ability to train large neural networks as policies for complex problems. In its standard form, TRPO only uses on-policy data, so its capacity for exploration is inherently limited. 

In experiments, we investigate three aspects of DE in comparison with baseline methods. First, the performance of {\it all} deployed policies through iterations of policy improvement. It is worth noting the importance of examining the performance of not only the main policy but also the perturbed policies in order to take the cost of exploration into account. Second, the pairwise KL divergence achieved by the perturbed policies of DE and RP, which measures the diversity of the perturbed policies. Third, the trace of the covariance matrix of perturbed gradient estimates. We demonstrate that high KL divergence correlates with a low trace of covariance in support of the theoretical analysis.  Additionally, we demonstrate the diminishing benefit of exploration when decreasing the number of perturbed policies.

\begin{figure*}[ht]
	\begin{tabular}{ccc}
		\includegraphics[width=0.28\textwidth]{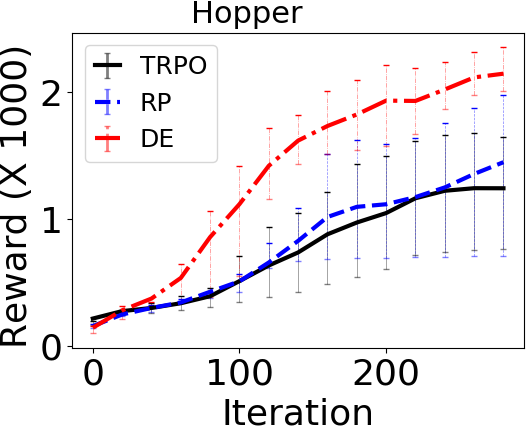}&
		\includegraphics[width=0.30\textwidth]{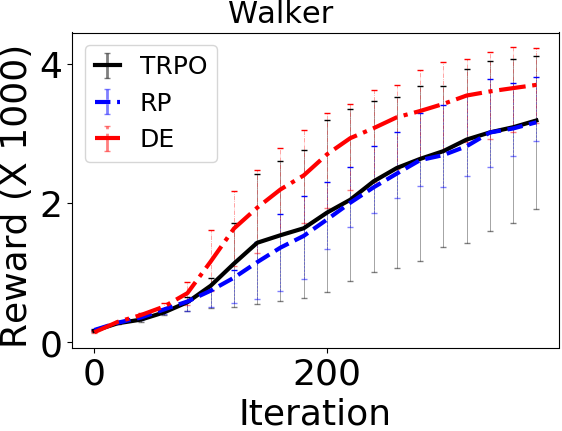}&
	    \includegraphics[width=0.30\textwidth]{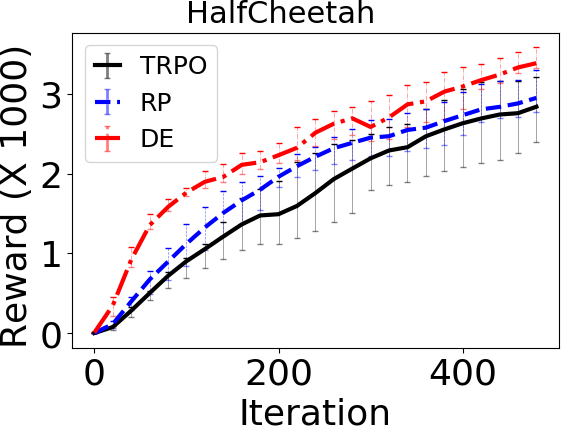}\\
        \vspace{5mm}
        (a)&(b)&(c)\\
        \includegraphics[width=0.325\textwidth]{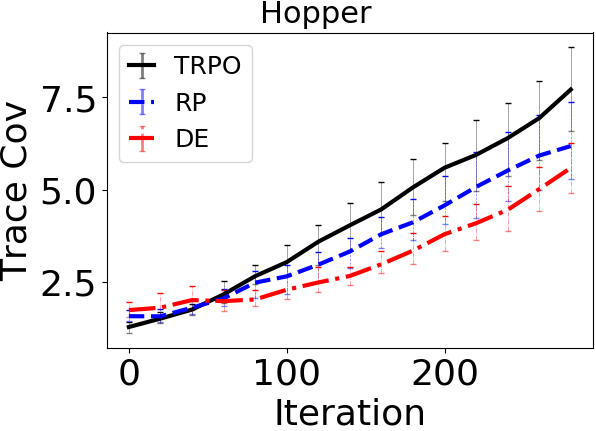}&
		\includegraphics[width=0.30\textwidth]{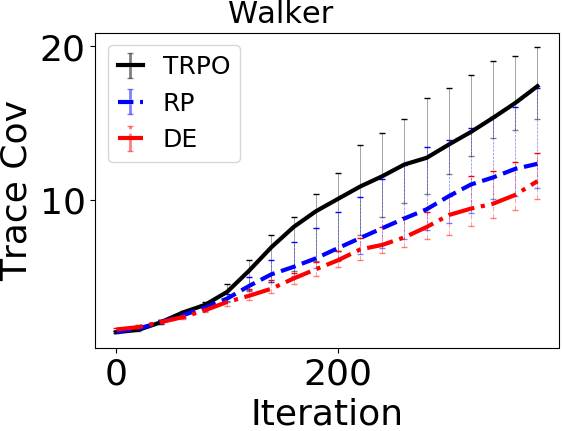}&
	    \includegraphics[width=0.30\textwidth]{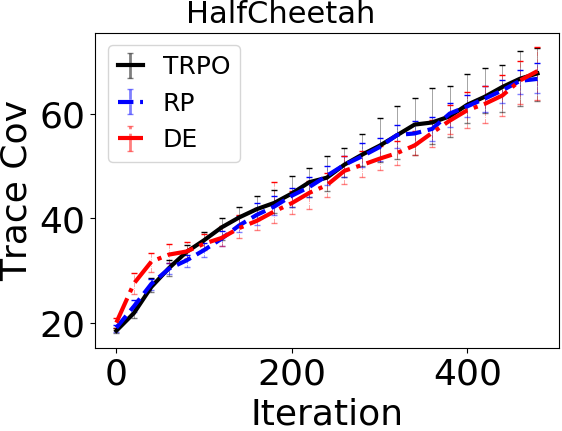}\\
        \vspace{5mm}
        (d)&(e)&(f)
	\end{tabular}
	\caption{Comparison between TRPO, RP (TRPO with Random Perturbations), and DE (TRPO with Diverse Exploration) on average performance of all behavior policies and trace of the covariance matrix of perturbed gradient estimates, across iterations of learning on (a,d) Hopper, (b,e) Walker and (c,f) HalfCheetah. Reported values are the average and interquartile range over 10 runs.}\label{fig:experiments}
	\label{experiments}
\end{figure*}

\subsection{Methods in Comparison}
We use two different versions of TRPO as baselines; the standard TRPO and TRPO with random perturbations (RP) and symmetric sampling. The RP baseline follows the same framework as DE but with random perturbations instead of conjugate perturbations. When implementing RP, we replace learning the covariance $\Sigma$ in the perturbed gradient estimate with a fixed $\sigma^2I$ as in \cite{PSNoise} in which it was noted that the computation for learning $\Sigma$ was prohibitively costly. The authors also propose a simple scheme to adjust $\sigma$ to control for parameter sensitivity to perturbations. The adjustment ensures perturbed policies maintain a bounded distance to the main policy. We achieve this by, for both conjugate and random, searching along the perturbation direction to find the parameterization furthest from the main policy but still within the perturbation radius $\delta_p$.
In light of the theoretical results, the use of symmetric sampling in RP serves as a more competitive baseline. 

Policies are represented by feedforward neural networks with two hidden layers containing 32 nodes and $tanh$ activation functions.  We found that increasing
complexity of the networks did not significantly impact performance and only increased computation cost. Additionally, we use layer normalization \cite{layernorm} as in \cite{PSNoise} to ensure that networks are sensitive to perturbations. Policies map a state to the mean of a Gaussian distribution with an independent variance for each action dimension that is independent of the state as in \cite{TRPO}. We significantly constrain the values of these variance parameters to align with the motivation for parameter perturbation approaches discussed in the Introduction.  This will also limit the degree of exploration as a result of noisy action selection. We use the TD(1) \cite{sutton} algorithm to estimate a value function $V$ over all trajectories collected by both the main and perturbed policies.
To estimate the advantage function, the empirical return of the trajectory is used as the $Q$ component and $V$ as a baseline.  TRPO hyperparameters are taken from \cite{TRPO,BenchmarkDeepRL}. 

We display results on three difficult continuous control tasks, Hopper, Walker and HalfCheetah implemented in OpenAI gym \cite{gym} and using the Mujoco physics simulator \cite{Mujoco}. As mentioned in the discussion of Algorithm~\ref{DPG}, the values of $k$, $\beta$ and $\beta_k$ determine exploration performed by perturbed policies. TRPO is at the extreme of minimal exploration since all samples come from the main policy. To promote exploration, in DE and RP we collect samples equally from all policies.  More specifically, we use $k=20$ perturbations for Hopper and $k=40$ perturbations for Walker and HalfCheetah for both DE and RP. Walker and HalfCheetah each have $3$ more action dimensions than Hopper and so require more exploration and hence more agents.   For a total of $N$ ($N=21000$ for Hopper and $N=41000$ for Walker and HalfCheetah in the reported results) samples collected in each policy improvement iteration, TRPO collects $\beta = N$ samples per iteration while DE and RP collect $\beta=\beta_k=\frac{N}{k+1}$ samples from the main and each perturbed policy. Through our experiments, we observed a trend of diminishing effect of exploration on policy performance when the total samples are held constant and $\beta$ increases. The initial perturbation radius used in experiments is $\delta_p=.2$ for Hopper and HalfCheetah and $\delta_p=.1$ for Walker.  Larger perturbation radiuses caused similar performance to the reported results but suffered from greater instability.  Reducing sensitivity to this hyperparameter is a direction for future research.

\subsection{Results}
\begin{table}
\centering
\caption{Total pairwise KL divergence averaged over iterations of DE vs. RP. Reported values are the average over 10 runs with all $p<0.001$.}
\begin{tabular}{|l|c|c|c|}
	\hline
	Domain & Hopper & Walker & HalfCheetah\\ 
	\hline	
	DE& 53.5 & 82.7 & 192.5\\
	\hline
	RP& 38.1 & 77.6 & 156.1\\
	\hline	
\end{tabular}
\label{KLtable}
\end{table}
The two rows of Figure~\ref{fig:experiments} and Table~\ref{KLtable} aim to address the three points of investigation raised at the beginning of this section.  Our goal is to show that perturbations with larger pairwise KL divergence are key to both strong online performance and enhanced exploration.  

In the first column of Figure~\ref{fig:experiments} and Table~\ref{KLtable}, we report results on the Hopper domain.  Figure (a) contains curves of the average performance (sum of all rewards per episode) attained by TRPO, RP and DE. For RP and DE, this average includes the main and perturbed policies.  RP has a slight performance advantage over TRPO throughout all iterations and converges to a superior policy. DE shows a statistically significant advantage in performance over RP and TRPO; a two-sided paired t-test of the average performance at each iteration yields $p<0.05$. Additionally, DE converges to a stronger policy and shows a larger rate of increase over both RP and TRPO. DE also results in the smallest variance in policy performance as shown by the interquartile range (IQR) which indicates that DE escapes local optima more consistently than the baselines. These results demonstrate the effect of enhanced exploration by DE over TRPO and RP.  

The trace of covariance of the perturbed gradient estimates are contained in Figure (d). Note, the covariance of TRPO gradient estimates can be computed by treating TRPO as RP but with policies perturbed by the zero vector. Interestingly, Figure (d) shows an increasing trend for all approaches.  We posit two possible explanations for this; that policies tend to become more deterministic across learning iterations as they improve and, for DE and RP, the decreasing perturbation radius.  Ultimately, both limit the variance of action selection and so yield more similar gradient estimates. Nevertheless, at any iteration, DE can significantly reduce the trace of covariance matrix due to its diversity. 

Column $1$ of Table~\ref{KLtable} reports the average total pairwise KL divergence over all perturbed policies for the Hopper domain. DE's conjugate policies have significantly larger pairwise KL divergence than RP.  This significant advantage in pairwise KL divergence yields lower variance gradient estimates which explain the observed superiority in performance, rate of improvement and lower IQR as discussed.

Similar trends are observed in Figures (b) and (e) and column $2$ in Table~\ref{KLtable} on the Walker domain. The performance of DE is clearly superior to both baselines but, due to the higher variance of the performance of the baselines, does not yield a statistically significant advantage. Despite this, DE maintains a significantly higher KL divergence between perturbed policies and significantly lower trace covariance estimates across iterations. Additionally, the same trends are observed in Figures (c) and (f) and column $3$ in Table~\ref{KLtable} in the HalfCheetah domain. DE shows a statistically significant advantage in terms of performance and pairwise KL divergence ($p<0.05$) over RP and TRPO despite their more similar covariance estimates.   

\begin{figure}
	\begin{tabular}{c}
		\includegraphics[width=0.4\textwidth]{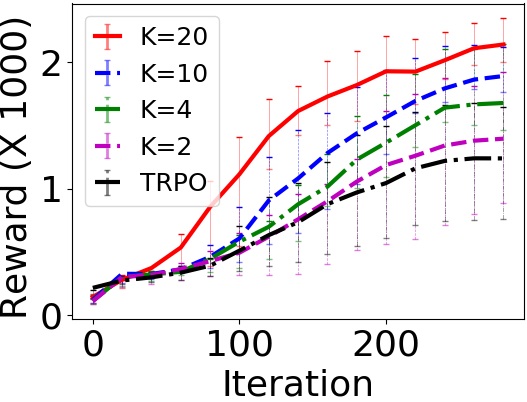}
	\end{tabular}
	\caption{Average performance of all behavior policies for DE on Hopper with a decreasing number of perturbed policies and TRPO.}\label{fig:ablation}
	\label{ablation}
\end{figure}

Finally, we present a study of the impact of decreasing the number of perturbed policies while keeping the samples collected constant on the Hopper domain.  In Figure~\ref{ablation}, we display the average performance of DE for $k=20,10,4,2$ as well as TRPO ($k=0$). Decreasing $k$ leads to decreasing average performance and rate of improvement. Additionally, decreasing $k$ leads to increasing performance variance.  Both of these observations demonstrate that increasing diversity among behavior policies is key to strong online performance and exploration.  

\section{Related Work}
Achieving exploration by deploying multiple policies has surfaced in the literature in varied contexts. The most closely related is \cite{PSNoise} which follows the same framework but does not provide theory or consider an optimal diversity objective.
\cite{DEFSI} provides a DE algorithm and theory for exploration under a safety model but does not address exploration for policy gradient methods.
\cite{Coord} uses multiple agents with diverse estimates of the MDP to efficiently learn a model of the reward and transition function. \cite{DDDRL} adds an explicit divergence regularizer to the policy gradient objective which encourages divergence from a heuristically chosen subset of previously deployed policies.

Others have studied parameter space noise for exploration in gradient-based evolutionary strategies
\cite{ESAltRL,PGPE,NES}, but they do not optimize diversity within policy performance constraints.
\cite{NoisyNets} proposes exploration by adding a trainable noise parameter to each network parameter, which incurs significant computational cost.

Different generalizations of the policy gradient with the common goal of variance reduction
in gradient estimates~\cite{expectedPG,interpolate} exist but do not address the same exploration issue studied in this work. An alternate line of work reduces variance using control variates \cite{ActDep}.

\section{Conclusions and Future Work}
We have proposed a novel exploration strategy and an algorithm framework for DE via conjugate policies for policy gradient methods. We have also provided a theoretical explanation for why DE works, and experimental results on three continuous control problems showing that DE outperforms the two baselines (TRPO and RP).

One future research direction is to investigate other efficient ways of generating a limited number of conjugate vectors that optimize KL divergence. Another is a more principled way of selecting the perturbation radius and reduction factor. More generally, future research will expand the DE theory and algorithm framework to other policy gradient methods and paradigms of RL.  

\section{Acknowledgements}
Yu's work is in part supported by the National Science Foundation (Award 1617915). Tong's work is in part supported by the National Nature Science Foundation of China (Award 61572418).

\bibliography{Cohen_etal}
\bibliographystyle{aaai}

\newpage
\appendix

\setcounter{thm}{0}
\section{Proof of Theorem 1}
\begin{thm}
Let $\epsilon_i$ and $\epsilon_j$ be two perturbations such that $\norm{\epsilon_i}_2$ = $\norm{\epsilon_j}_2=\delta_\epsilon$. Then, (1) the trace of $\textrm{Cov}(\nabla_{\phi}\log(\pi_{\epsilon_j}),\nabla_{\phi}\log(\pi_{\epsilon_i}))$ is minimized and (2) $\frac{1}{2}(\epsilon_j-\epsilon_i)^T\hat F(\epsilon_i)(\epsilon_j-\epsilon_i)$ the estimated KL divergence $D_{KL}(\pi_{\epsilon_i}||\pi_{\epsilon_j})$ is maximized, when $\epsilon_i=-\epsilon_j$ and they are along the direction of the eigenvector of $F(\epsilon_i)$ with the largest eigenvalue.
\begin{proof}

First, we note that the covariance
\begin{align*}
&\mathrm{Cov}[\nabla_\phi\log(\pi_{\epsilon_j}),\nabla_\phi\log(\pi_{\epsilon_i})]\\
=&\mathbb{E}_{s,a}[\nabla_\phi\log(\pi_{\epsilon_j}(a|s))\nabla_\phi\log(\pi_{\epsilon_i}(a|s))^T]\\
&-\mathbb{E}_{s,a}[\nabla_\phi\log(\pi_{\epsilon_i}(a|s))]^T\mathbb{E}_{s,a}[\nabla_\phi\log(\pi_{\epsilon_j}(a|s))]\\
=&\mathbb{E}_{s,a}[\nabla_\phi\log(\pi_{\epsilon_j}(a|s))\nabla_\phi\log(\pi_{\epsilon_i}(a|s))^T]
\end{align*}
since the score function $\mathbb{E}_{s,a}[\nabla_\phi\log(\pi_{\epsilon_j}(a|s))] = 0$.

The arguments $s,a$ are dropped in the following for clarity. Consider the multivariable Taylor expansion,
\begin{align*}
    \nabla_\phi\log(\pi_{\epsilon_j}) = \nabla_\phi\log(\pi_{\epsilon_i}) + \nabla^2_\phi\log(\pi_{\epsilon_i})(\epsilon_j-\epsilon_i) + R(\pi_{\epsilon_j}),
\end{align*}
where the remainder $R(\pi_{\epsilon_j})$ is a vector whose $k$th element is $(\epsilon_j-\epsilon_i)^T\nabla^2_\phi(\nabla_{\phi_k}\log(\pi^*))(\epsilon_j-\epsilon_i)$, in which $\nabla_{\phi_k}$ represents the partial derivative with respect to $\phi_k$ the $k$th component of $\phi$ and $\pi^* =\pi_{\phi+\epsilon^*}$ and $\epsilon^* = \epsilon_i + c(\epsilon_j-\epsilon_i)$  for $0<c<1$. Taking the trace of the covariance, we have
\begin{align*}
&\textrm{Trace}\{\mathbb{E}_{s,a}[\nabla_\phi\log(\pi_{\epsilon_j})\nabla_\phi\log(\pi_{\epsilon_i})^T]\}\\
=&\textrm{Trace}\{\mathbb{E}_{s,a}[\nabla_\phi\log(\pi_{\epsilon_i})\nabla_\phi\log(\pi_{\epsilon_i})^T\\
&+ \left[\nabla^2_\phi\log(\pi_{\epsilon_i})(\epsilon_j-\epsilon_i)\right]\nabla_\phi\log(\pi_{\epsilon_i})^T+ R(\pi_{\epsilon_j})\nabla_\phi\log(\pi_{\epsilon_i})^T]\}\\
=&\textrm{Trace}\{ F(\epsilon_i)\} -\textrm{Trace}\{\mathbb{E}_{s,a}[\hat F(\epsilon_i)(\epsilon_j-\epsilon_i)\nabla_\phi\log(\pi_{\epsilon_i})^T\}]\\
 &+o(\|\epsilon_j-\epsilon_i\|)\\
=&\textrm{Trace}\{ F(\epsilon_i)\} -\{\mathbb{E}_{s,a}[\nabla_\phi\log(\pi_{\epsilon_i})^T\hat F(\epsilon_i)(\epsilon_j-\epsilon_i)]\}\\
&+o(\|\epsilon_j-\epsilon_i\|)
\end{align*}
Since the first term is independent of the choice of $\epsilon_i$ and $\epsilon_j$ and the remainder term converges to zero as $\delta\rightarrow 0$, to minimize the trace of the covariance we need to focus on maximizing the second term.
\begin{align*}
&\mathbb{E}_{s,a}[\nabla_\phi\log(\pi_{\epsilon_i})^T\hat F(\epsilon_i)(\epsilon_j-\epsilon_i)]\\
=&\mathbb{E}_{s,a}\{[(\epsilon_j-\epsilon_i)^T\hat F(\epsilon_i)\nabla_\phi\log(\pi_{\epsilon_i})
\nabla_\phi\log(\pi_{\epsilon_i})^T\hat F(\epsilon_i)(\epsilon_j-\epsilon_i)]\}^{1/2}\\
=&\mathbb{E}_{s,a}\{(\epsilon_j-\epsilon_i)^T\hat F(\epsilon_i)\tilde F(\epsilon_i)\hat F(\epsilon_i)(\epsilon_j-\epsilon_i)\}^{1/2}\\
\approx&\{(\epsilon_j-\epsilon_i)^TF(\epsilon_i)^3(\epsilon_j-\epsilon_i)\}^{1/2},
\end{align*}
Here $F(\epsilon_i)$ is the true Fisher information, $\hat F(\epsilon_i)$ is the observed information, and $\tilde F(\epsilon_i)=\nabla_\phi\log(\pi_{\epsilon_i})
\nabla_\phi\log(\pi_{\epsilon_i})^T$, and we note that $\mathbb{E}(\hat F(\epsilon_i))=\mathbb{E}(\tilde F(\epsilon_i))=F(\epsilon_i)$. Consider the eigen-decomposition of $F(\epsilon_i)=U\Lambda U^T$. We have

\begin{align*}
&(\epsilon_j-\epsilon_i)^TF(\epsilon_i)^3(\epsilon_j-\epsilon_i)\\
=&(\epsilon_j-\epsilon_i)^TU\Lambda^3 U^T(\epsilon_j-\epsilon_i)\\
=&\sum_{k=1}^p \lambda_k^3 [u_k^T(\epsilon_j-\epsilon_i)]^2.
\end{align*}
This objective is maximized at $\epsilon_j=-\epsilon_i=\delta u_1$ (or $-\delta u_1$), that is, they are opposite to one another and they are along the eigenvector corresponding to the largest eigenvalues $\lambda_1$. In this case, the maximal value is $4\delta^2\lambda_1^3$.

We now note that $(\epsilon_j-\epsilon_i)^TF(\epsilon_i)^3(\epsilon_j-\epsilon_i)$ is closely related to the KL-divergence $\frac{1}{2}(\epsilon_j-\epsilon_i)^T F(\epsilon_i)(\epsilon_j-\epsilon_i)$. Using the same argument as above, this quantity is maximized by $\epsilon_j=-\epsilon_i=\delta u_1$ or ($-\delta u_1$).

In conclusion, the pair of pertubations $\epsilon_j=-\epsilon_i=\delta u_1$ maximize the KL divergence and minimize the trace of the covariance $\textrm{Trace}\{\mathbb{E}[\nabla_\phi\log(\pi_{\epsilon_j})\nabla_\phi\log(\pi_{\epsilon_i})^T]\}$ up to a constant which converges to 0 faster than $\delta\rightarrow 0$.
\end{proof}
\end{thm}

\section{Proof of Theorem 2}
\begin{thm}
The set of conjugate vectors $\{\mu_1, \mu_2,\ .\ .\ ,\mu_k\}$ maximize the objective~(\ref{objf}) among any $k$ perturbations.
\begin{proof}
Let $\mathcal{P} = \{\epsilon_i\ |\ 2 \le i \le k \}$, where $\epsilon_i = \eta_{i1}\mu_1 + \eta_{i2}\mu_2 +\ .\ .\ +\eta_{in}\mu_n$, be a set of $k$  perturbations. First, we show the result for $k=2$. In this case, the value of the objective~(\ref{objf}) is
\begin{align*}
&\tilde{D}_{KL}(\epsilon_1||\epsilon_2)\\
=&\sum_{i=1}^n (\eta_{1i}-\eta_{2i})^2\norm{\mu_i}_F\\
=&\sum_{i=1}^n (\eta_{1i}^2-2\eta_{1i}\eta_{2i}+\eta_{2i}^2)\norm{\mu_i}_F\\
=&\sum_{i=1}^n \eta_{1i}^2\norm{\mu_i}_F+\sum_{i=1}^n \eta_{2i}^2\norm{\mu_i}_F-2\sum_{i=1}^n \eta_{1i}\eta_{2i}\norm{\mu_i}_F
\end{align*}
Recall that $\eta_{1i},\eta_{2i}\ge 0$. Therefore, to maximize the objective, we must select $\epsilon_1$ and $\epsilon_2$ so that $\eta_{1i}\eta_{2i}=0$ for all $i$. Next, since $\norm{\mu_1}_F\ge \norm{\mu_2}_F\ge \dots$, it suffices to let $(\eta_{11}, \eta_{12}, \dots\eta_{1n}) =(1,0,\dots,0)$ and  $(\eta_{21}, \eta_{22},\eta_{23} \dots\eta_{2n}) =(0,1,0,\dots,0)$.

The same argument can be generalized to $k>2$. In the end, we choose $(\eta_{s1}, \eta_{s2}, \dots\eta_{sn})$ to be the vector of all zeros except 1 at the $s$th entry. This choice corresponds to the case that the $s$th perturbation is the $s$th conjugate vector.
\end{proof}
\end{thm}

\end{document}